\documentclass[pdflatex,sn-mathphys-ay]{sn-jnl}

\usepackage{graphicx}%
\usepackage{multirow}%
\usepackage{amsmath,amssymb,amsfonts}%
\usepackage{amsthm}%
\usepackage{mathrsfs}%
\usepackage[title]{appendix}%
\usepackage{xcolor}%
\usepackage{textcomp}%
\usepackage{manyfoot}%
\usepackage{booktabs}%
\usepackage{algorithm}%
\usepackage{algorithmicx}%
\usepackage{algpseudocode}%
\usepackage{listings}

\theoremstyle{thmstyleone}%
\theoremstyle{thmstyletwo}%

\theoremstyle{thmstylethree}%
\newtheorem{definition}{Definition}%

\newcommand{\BibTeX}{B\kern-.05em{\sc i\kern-.025em b}\kern-.08em\TeX}

\definecolor{orange}{RGB}{255,127,0}
\definecolor{brown}{RGB}{150,70,0}
\definecolor{green}{RGB}{127,255,127}
\definecolor{darkgreen}{RGB}{0,127,0}
\definecolor{blue}{RGB}{127,127,255}
\definecolor{lightblue}{RGB}{150,150,255}
\definecolor{darkblue}{RGB}{0,0,127}
\definecolor{red}{RGB}{255,90,90}
\definecolor{grey}{RGB}{127,127,127}
\definecolor{pink}{RGB}{255,180,180}
\definecolor{fuchsia}{RGB}{255,0,255}

\begin{document}

\title[Article Title]{Evaluating Simplification Algorithms for Interpretability of Time Series Classification}


\author*[1]{\fnm{Brigt} \sur{Håvardstun}}\email{brigt.havardstun@uib.no}

\author[2]{\fnm{Félix} \sur{Martí-Perez}}\email{fmarper@inf.upv.es}

\author*[2]{\fnm{Cèsar} \sur{Ferri}}\email{cferri@dsic.upv.es}
\author[1]{\fnm{Jan Arne} \sur{Telle}}\email{Jan.Arne.Telle@uib.no}

\affil*[1]{\orgdiv{Department of Informatics}, \orgname{University of Bergen}, \orgaddress{ 
\city{Bergen},
\country{Norway}}}

\affil[2]{\orgdiv{VRAIN}, \orgname{Universitat Politècnica de València}, \orgaddress{
\city{València}, 
\country{Sapin}}}



\abstract{
In this work, we introduce metrics to evaluate the use of simplified time series in the context of interpretability of a TSC - a Time Series Classifier.
Such simplifications are important because time series data, in contrast to text and image data, are not intuitively understandable to humans.
These metrics are related to the \emph{complexity} of the simplifications - how many segments they contain - and to their \emph{loyalty} - how likely they are to maintain the classification of the original time series. We focus on simplifications that select a subset of the original data points, and show that these typically have high Shapley value, thereby aiding interpretability.
We employ these metrics to experimentally evaluate four distinct simplification algorithms, across several TSC algorithms and across datasets of varying characteristics, from seasonal or stationary to short or long. 
We subsequently
perform a human-grounded evaluation with forward simulation, 
that confirms also the practical utility of the introduced metrics to evaluate the use of simplifications in the context of interpretability of TSC. Our findings are summarized in a framework for deciding, for a given TSC, if the various simplifications are likely to aid in its interpretability. 
}

\keywords{Interpretability, Time Series Classification, Simplifications, User Evaluations}



\maketitle

\section{Introduction}

Temporal data is encountered in many real-world applications ranging from patient data in healthcare \citep{Rajkomar} to the field of cyber security \citep{Susto}. Deep learning methods have been successful for TSC - Time Series Classification -   \citep{Fawaz,Rajkomar,Susto} but such methods are not easily interpretable, and often viewed as black boxes, which limits their applications when user trust in the decision process is crucial. To enable the analysis of these black-box models we revert to post-hoc interpretability.
Recent research has focused on adapting existing methods to time series, where we can find specific methods like SHAP-LIME \citep{Guilleme}, Saliency Methods \citep{Ismail}, and Counterfactuals \citep{Delaney}, and also combinations of these \citep{Schlegel}. 

As humans learn and reason by forming mental representations of concepts based on examples, and any machine learning model has been trained on data, then we believe that data e.g. in the form of prototypes and counterfactuals is indeed the natural common language between the user and this model. However, the basic problem is that compared to images and text, time series data are not intuitively understandable to humans \citep{Siddi}. 
This makes interpretability of time series extra demanding, both when it comes to understanding how users will react to the provided explanations and to predict what explanatory tools are best.
For example, in \citep{DemoECML} a tool was given for explainability of a TSC, that allowed model inspection so the user could form their own mental model of the classifier. However, a user evaluation showed that non-expert end-users were not able to make use of this freedom, 
supporting the notion that time-series are particularly non-intuitive for humans. 

In \citep{Theissler}, authors give an intriguing taxonomy of XAI for TSC, divided into those focused on (i) Time-points (e.g. SHAP and LIME) or (ii) Subsequences (e.g. Shapelets) or (iii) Instances (Prototypes, CF, Feature-based). Explaining a classifier by instances like prototypes has a definite appeal, but it is a challenge how to highlight features important for the classification, while at the same time simplifying the instance to mitigate the non-intuitive aspects of this domain.
In \citep{Theissler} the authors also review the literature on XAI for TSC and of the 9 papers they discuss on Instance-based explanations using Prototypes or Features, it is remarkable that not a single one is specialised for Local explanations, rather they are all Global.
The algorithms we evaluate in this paper fill this gap, as they are ways of \emph{simplifying} a time series by straight-line segments\footnote{Piecing together several such local explanations, say for prototypes of each class, this can still form a global explanation of the given TSC.}.
We focus on Piecewise-linear sketches since they keep every retained point on the true time-and-value axes, so users can read magnitudes and slopes directly, without translating bins or symbols. This matches the canonical line-chart format that dashboards and decades of perception studies show that people understand quickly and accurately \citep{cleveland1984graphical,heer2010crowdsourcing}.

We introduce metrics to evaluate the use of algorithms computing simplified time series in the context of interpretability of a TSC. This  addresses the key challenge that time series data, in contrast to text and image data, are not intuitively understandable to humans. 
These metrics are related to the \emph{complexity} of the simplifications produced by the algorithms - how many data points they maintain as a percentage of the original - and to their \emph{loyalty} - how likely they are to maintain the classification of the original time series.
We first employ these metrics to theoretically and experimentally evaluate four distinct simplification algorithms, across several TSC algorithms and across datasets of varying characteristics, from seasonal or stationary to short or long. We also employ the metrics in a practical setting, doing a quantitative user evaluation, thereby meeting a demand from
the XAI research community \citep{Delaney, Theissler, wang2023gam}, to measure how simplifications increases the understanding that the user has of the black
box classification. Using the taxonomy of interpretability evaluation from \citep{doshi2017towards}, what we
do is a Human-Grounded Evaluation with Forward Simulation, to evaluate if simplifications are better than full time series, when using prototypes to teach the classification used by a TSC.

Our findings suggest that using simplifications for interpretability of TSC is in some cases much better than using the original time series. However, we discover various cases where the trade-off between complexity and loyalty, or some other characteristics of the given dataset and TSC, are such that simplifications are not very useful. 
This happens for example when the original time series have several data points lying on an almost straight line, in which case the human eye seems to perform the simplification automatically, and the user interprets the TSC equally well from original time series as from loyal simplifications. In short, our findings suggest that simplifications can be useful when i) the TSC is not itself easily interpretable on the original domain, ii) the simplifications are loyal but not complex and dissimilar to the original, and moreover iii) the simplified domain is understandable under labels from TSC on the original domain. 

In the rest of this paper we first discuss related work and cover some standard definitions before presenting the metrics we will use, the four simplification algorithms we will evaluate, the TSCs we will employ and the 40 UCR datasets we use with their varying characteristics.
We then show the results of our theoretical experiments, first comparing the four algorithms over a variety of dataset characteristics, and then focusing on the best performing algorithms and evaluating how useful they may be for interpretability. We then report on the user study, and based on the lessons learned we give a framework for deciding if simplifications are likely to aid in interpretability of a given TSC.

\section{Related Work}

Several techniques have been applied to generate explanations from TSC models \citep{Theissler,rojat2021explainable}. Specifically, techniques previously used on Convolutional Neural Networks or Recurrent Neural Networks have been applied for TSC. For instance, 
the authors in \citep{Schlegel} apply to time series several XAI methods previously used on image and text domain. They also  introduce verification techniques specific to times series, in particular a perturbation analysis and a sequence evaluation.  

Another alternative is to produce explanations through examples, and these can be specifically utilised in the time series domain. One type of this explanation method involves giving the nearest example from the training dataset that acts as a prototype to depict the normal behaviour of a similar sample \citep{geler2020weighted}. 
A method to generate prototypes for time series data using an autoencoder is presented in \citep{Gee}. The main novelty of the work is the method to generate diverse prototypes. 


Given that in many cases raw time series can be too complex for humans, several studies have tried to employ simplified versions as explanations. 
In \citep{keogh2001dimensionality}, the authors propose a dimensionality reduction technique for time series, called Piecewise Aggregate Approximation, as a tool for similarity search in large time series databases. 
In \citep{keogh2004segmenting}, a comprehensive survey on time series segmentation, the authors highlight that the optimal number of segments often depends on the underlying patterns in the data and the specific objectives of the analysis.
Also,  in \citep{camponogara2015models}, the authors introduce a range of piecewise-linear models and algorithms for unknown functions. Another option is to segment time series into fixed-width windows and employ these to justify decisions. In \citep{schlegel2021tsmule}, the authors propose TS-MULE, a model explanation method working to segment and perturb time series data and extending LIME. A study on the effect of segmentation on time series, in particular in the field of finance for the use of pattern matching was presented in \citep{WAN2016346}. 
In \citep{SI20132581}, the researchers also apply segmentation to financial time series data, now as a preprocessing step to locate technical patterns. Similarly, in \citep{wu2021optimal}, the authors employ financial time series segmentation and introduce the Optimal Multi-Segment Linear Regression (OMSLR) algorithm.  Their approach aims to minimise the global mean square error (between the simplification and the original time series ) for a given number of segments $k$.

\section{Simplifications and Metrics}
We first recall basic notions and then introduce the simplifications and metrics that we will study in this paper.

Staying consistent with earlier notation \citep{Theissler,Delaney} a time series $T=\{t_1,t_2,…,t_n\}$ is an ordered set of $n$ real-valued observations (or time steps). Note we may also view each $t_i=(x_i,y_i)$ as a pair consisting of an $x$-value (the time) and a $y$-value (the observation), where $x_i < x_{i+1}$, and often $x_i=i$.
A time series dataset $D=\{(T_1, c_1), (T_2, c_2),...,(T_m, c_m)\}$  is a collection of such time series where each time series $T_i\in R^{m}$ has a class label $c_i$. Thus, for binary classification tasks we would have $c_i \in \{0,1\}$.
Given such a dataset $D$, Time Series Classification (TSC) is the task of training a mapping $C_D$ from the space of possible inputs to a 
predicted class. Interpretability of TSC concerns giving a human user an understanding of how such a trained TSC decides on a predicted class. Exemplar-based interpretability employs showing examples of classifications made by the TSC, and this will allow the user to form their own mental model of the classifier.
Prototypes are time series exemplifying the main aspects responsible for a classifier’s specific decision outcome. Showing prototypes is well-known in exemplar-based interpretability. 

For TSC interpretability, in particular for long time series, such prototypes often have too much information, and we therefore focus on what we call simplifications. 
Such simplifications have been common practice in data mining and other related fields, and hence a wide range of simplification techniques have been constructed to suit different needs. In our work we focus on trustworthy and faithful simplifications. In the Experiments and Results section, we empirically show that simplifications that select a subset of original time steps as their simplifications are more faithful to important time steps.



\begin{figure}[h]
\centering
\includegraphics[width=10cm]{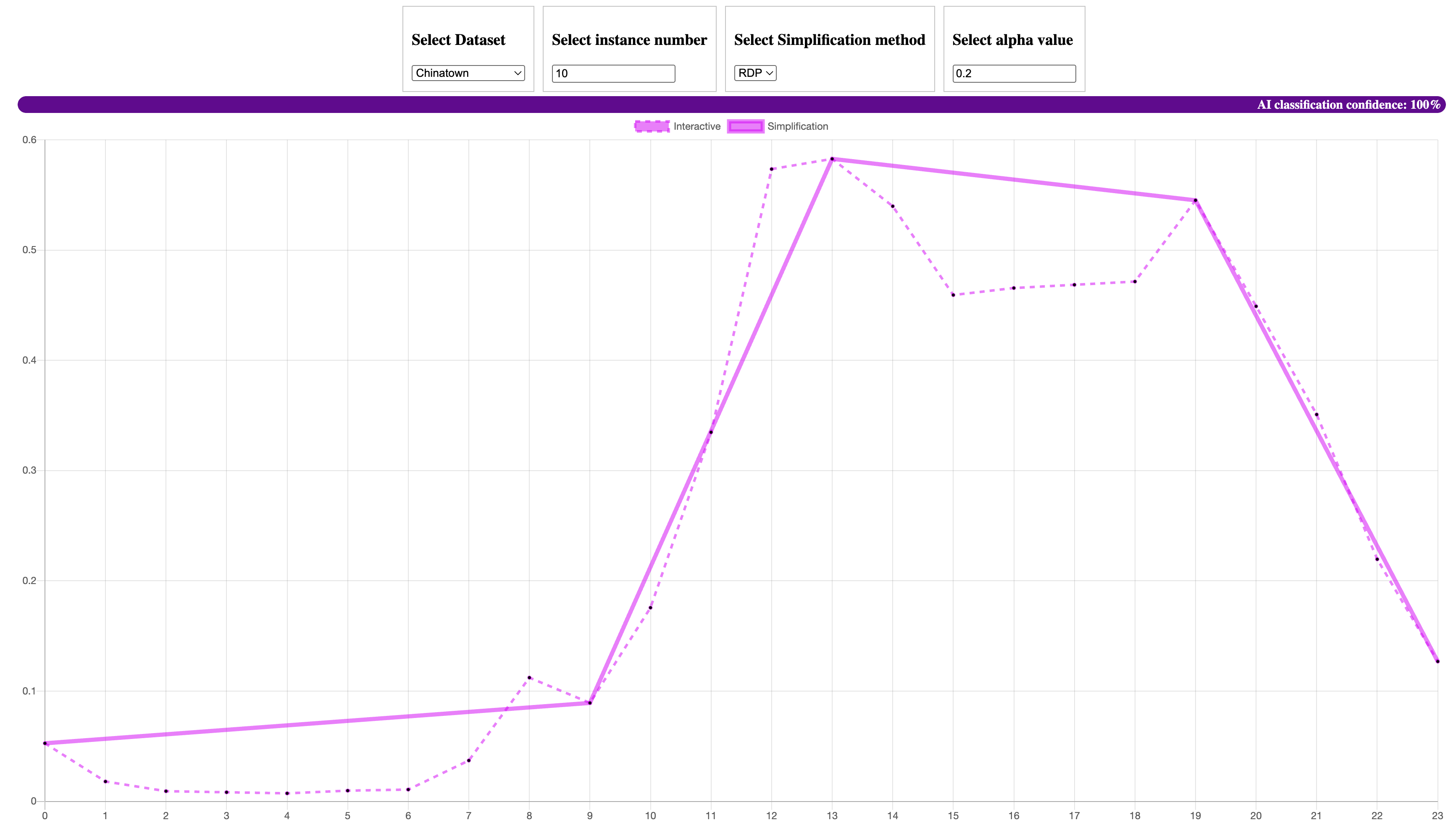}
\caption{An original time series (dotted line) and the simplification on 4 segments produced by RDP with parameter $\alpha=0.2$ (used to set the level of simplification)}.
\label{fig:tool}
\vspace{-0.3cm}
\end{figure}

Hence for the purposes of this paper a \emph{simplification} of a time series $ts=\{t_1,t_2,…,t_n\}$ on $|ts|=n$ time steps consists of dropping some of the time steps, thus keeping only time steps at a subset $S \subseteq \{1,2,...,n\}$ of time points and replacing the rest by the points lying on the straight-line segments given by the kept points, thus resulting in a new time series $sts=\{t'_1, t'_2,...,t'_n\}$ with $t'_i=t_i=(x_i,y_i)$ for $i \in S$ and otherwise $t'_i=(x_i,y'_i)$ with $y'_i$ the value at $x_i$ of the straight line between $t_j$ and $t_k$ for $j < k$ and $\{j,j+1,...,k\} \cap S=\{j,k\}$. We will say that the simplification $sts$ consists of $|S|-1$ segments, and we denote its number of segments by $||sts||$. The selection strategy of the specific subset of time steps depends on the algorithm and  parameters used. See in Figure \ref{fig:tool} an example showing a simplification with 4 segments obtained by keeping 5 time steps (at times 0,9,13,19,23) of an original time series on 24 time steps (at times 0 to 23). 

A simplification algorithm for time series called $A$ takes as input an original time series $ts$ and outputs a simplification $A(ts)$ on $||A(ts)||$ segments. Note that the length of $A(ts)$ is the same as the length of $ts$ so a classifier trained for time series of that length can be used on the simplification $A(ts)$.

\begin{definition}
    To evaluate the usefulness of a simplification algorithm $A$  for interpretability of a TSC $C_D$, we introduce the following metrics. Let $P$ be a representative subset of inputs, chosen randomly according to the expected distribution of inputs to $C_D$, and let these original inputs have length $n$:

\begin{itemize}
\item Complexity: : $\frac{1}{n}\frac{1}{|P|}\sum_{ts \in P} ||A(ts)||+1$, the expected number of time points kept in the simplification, as a fraction of the length of the original 
\item Loyalty: $\frac{1}{|P|}|\{ts \in P: C_D(ts)=C_D(A(ts)) \}|$, the expected probability that the simplification has the same classification as the original
\item Kappa loyalty: $\frac{p_0-p_e}{1-p_e}$ where $p_0=\frac{1}{|P|}\sum_{ts \in P}1[C_D(ts)=C_D(A(ts))]$ is the relative observed agreement, and $p_e=\sum_{c \in C} P_{CD}(c) \cdot P_{CD(AC)}(c)$ is the hypothetical probability of chance agreement
\end{itemize}
\end{definition}

Thus complexity is a value between $\frac{2}{n}$ (when all simplifications of instances in $P$ are on a single segment) and $1$ (when all simplifications are the trivial one which does not remove any time step from the original), while loyalty is a value between 0 (when all simplifications are classified by $C_D$ as different from the original ones) and 1 (when complete agreement, all simplifications are classified the same as the original ones). The Kappa loyalty \citep{Coh60} will have a maximum value of 1 when we have complete agreement, a value of 0 if the agreement is no better than chance, and below 0 if a tendency for different classes.
Kappa was used as a performance metric as it accounts for agreement occurring by chance, providing a more robust measure than accuracy, particularly in the presence of class imbalance.
Note that when using a simplification algorithm for interpretability of TSC, showing the simplifications to a user, then low complexity should allow a more easy focus on important features. Regarding loyalty, let us first emphasize that what we intend to teach the user are the patterns discovered by the TSC in classifying original time series. If we show the user simplified time series with Kappa loyalty 1 then nothing has been lost, since whatever the user learns from the simplifications carries over to the original. Note that we can expect this to happen for some complexity below 1, when all such simplifications retain the classification of the originals. However, if we show the user simplifications with loyalty $\lambda < 1$ then what the user learns does not fully carry over to the original, and we will take this into account when evaluating its utility. When applying simplifications we are thus searching for a sweet-spot between high loyalty and low complexity, to maximize the users ability to learn the patterns actually used by the TSC.

\section{Simplification Algorithms}
We briefly describe the four simplification algorithms we have chosen to focus on. Although they are quite distinct, they share a number of properties, which will allow for a fair comparison:

\begin{itemize}
\item their input consists of an original time series, given by $n$ points 
\item their output is a simplification given by a subset of input points 
    \item they have a parameter $\alpha$ and increasing its value will monotonically decrease number of segments in output, from $n-1$ to 1.
\end{itemize}

\subsection{Ramer-Douglas-Peuker}

RDP is a recursive algorithm introduced in \citep{RAMER1972244, doi:10.3138/FM57-6770-U75U-7727}. Given $n$ points $p_1,...,p_n$ select a point $p_d$ with the largest distance to the straight line between the first point $p_1$ and
last point $p_n$. If the distance to $p_d$ is no more than $\alpha$, then we are in a
base case and return the two points $p_1$ and $p_n$. If the distance to $p_d$
exceeds $\alpha$, then recursively process the two sequences of points
formed by $p_1,...,p_d$ and $p_d,...,p_n$. 
The implementation (\citep{rdp_0.8}) used for RDP has runtime $O(n \log n)$.

\subsection{Visvalingam–Whyatt algorithm}

 VW is an iterative algorithm introduced in \citep{VW}. It computes areas of triangles formed by
successive triples of points $a,b,c$; in each iteration, the middle point $b$ with the
smallest associated triangle is removed, the area of neighbouring
triangles is recomputed, and the process is repeated.
The process stops when the area of the smallest triangle exceeds $\alpha$. 
The implementation of  VW we use \citep{Hugel_Simplification_2021}  has runtime $O(n \log n)$.

\subsection{Bottom-Up}

BU is based on an iterative algorithm as described in \citep{Keogh} with the same name. 
We slightly modified the algorithm from \citep{Keogh} so that it adheres to the simplification algorithms we study here - selecting subsets of original time steps. Initially each time point is viewed as a trivial seg (a special segment), and in each iteration we merge two consecutive segs A and B (if A ends in time step $k$ then B starts in time step $k+1$) with the merged seg AB represented by a straight line going from the start of A to the end of B (in original BU the segment AB would be represented by the Least Square Fit over this area). The two segs merged into some AB in any iteration should have lowest error, which is the sum over each time point $t$ within AB, of the absolute value of the difference between the value of the original time series at $t$ and the value of the AB line at $t$. We stop when the lowest error exceeds $\alpha$, with $s$ segs $S_1,S_2,...,S_s$ such that if $S_i$ ends in $k$ then $S_{i+1}$ starts in $k+1$, and we return the corresponding simplified time series of length $2s$. While the resulting simplification has only $s$ segs, if we add the gaps between consecutive segs we get $2s -1$ ordinary segments. 
Our implementation of BU has runtime $O(n \log n)$.

\subsection{Optimal Simplification}
OS is a dynamic programming algorithm  introduced in \citep{ORS}. It finds the piecewise linear simplification (extending the first and last of its segments to the start and end respectively) that minimizes the sum of: Squared Euclidean Distance to the original time series plus $\alpha$ times the number of segments used. 
Our implementation of OS has runtime $O(n^3)$. Note that, in contrast to the other three algorithms, it finds the optimal simplification for a given $\alpha$, rather than just being a greedy algorithm, however this leads to it being more time consuming.


\subsection{Normalized complexity parameter}

To ensure uniformity we alter each of the four simplification algorithms so that it has a normalized complexity parameter called
$\alpha_c$. This parameter will range between a minimum of 0 (corresponding to a high value of $\alpha$ giving a small number of segments as output) and a maximum of 1 (corresponding to a low value of $\alpha$ giving a high number of segments as output).

\section{Datasets and Classifiers}
\subsection{Datasets Selection}

The UCR repository \citep{UCR} contains 128 datasets for univariate TSC. Their features cover problems from binary up to 52-class classification, with time series lengths ranging from 15 to 2,700 data points. Given that the datasets come from a wide range of domains, such as sensors, medical devices, or traffic, the characteristics of these datasets are diverse.
Most datasets come with predefined splits for training, and test sets. 

To keep runtime manageable, we included only UCR datasets with time series of length less than 200 data points. This has resulted in 40 datasets of varying length, class and domain. See Table \ref{tab:big} for details on the 40 chosen datasets. 
Additionally, we categorize each dataset by stationarity, seasonality, and entropy, enabling in‐depth analysis of how different simplification algorithms perform under varying data characteristics.

\subsection{Feature Characterisation}

The Augmented Dickey Fuller (ADF) test \citep{adf} is a statistical test used to determine the presence of a unit root in the series, thus determining if a time series is non-stationary. 
We set the significance level at 0.05 and apply the test to each instance: if the p-value is below it, we reject the null hypothesis that a unit root is present in the time series and label that instance “stationary”; otherwise, we label it “non-stationary.” At the dataset level, we declare a dataset “stationary” if at least 80\% of its instances are stationary (supermajority), “non-stationary” if at most 50\% are stationary (simple-majority), and “partially-stationary” when the proportion lies between these thresholds \citep{WikipediaSupermajority}.

The autocorrelation function (ACF) measures the correlation of a time series with a lagged version of itself. We utilise an extension that uses Fast Fourier transform (FFT) convolution to make ACF faster and more suitable for longer time series. A dataset is considered to be seasonal if most of its instances have a correlation coefficient greater than 0.4, thus being seasonal. This cut-off was decided empirically by visually inspecting the time series. 

Approximate entropy \citep{doi:10.1073/pnas.88.6.2297} is a technique used to quantify the unpredictability of fluctuations in time series. This returns a continuous value between 0 (low entropy) and 1 (high entropy), which will determine the entropy of the time series. To obtain the entropy of a dataset, we obtain the mean entropy over all instances. 
To facilitate the analysis, we discretize the entropy scores into three equal-frequency bins. We determine the breakpoints to be 0.23 and 0.33. Instances with entropy up to 0.23 are labelled as low entropy, those between 0.23 and 0.33 as medium entropy, and those above 0.33 as high entropy.

\subsection{Classifiers}
\label{classifiers}

Different types of classification algorithms are used for TSC. Throughout the literature, the most common algorithms are:
\begin{itemize}
    \item Logistic Regression: A linear model that estimates the probability of each class using weighted input features; it is simple, interpretable (white-box), and best suited for linearly separable time series. We use scikit-learn’s LogisticRegression \citep{scikit-learn} with its default L2 penalty and solver.
    \item Decision Trees: A non-linear, rule-based model (white-box) that recursively partitions the feature space to make decisions. Capable of handling complex relationships and capturing non-linearities in time series, but prone to overfitting. Implemented via scikit-learn’s DecisionTreeClassifier \citep{scikit-learn}  using the Gini impurity criterion and no maximum-depth constraint.
    \item k-Nearest Neighbour (kNN): A non-parametric method that classifies a time series based on the majority label among its closest examples in the training set. It can be interpretable (white-box), but its decisions can become less transparent with higher dimensionality. We leverage tslearn’s KNeighborsTimeSeriesClassifier \citep{JMLR:v21:20-091}, configured with k = 5 and distance-weighted voting.
    \item Convolutional Neural Networks (CNN): A deep learning architecture that automatically extracts and learns hierarchical features through convolutional filters is highly effective for discovering patterns in time series data. However, it is considered a black-box due to its complex and opaque decision processes.  Our networks are built in PyTorch 
    consisting of three 1D convolutional layers, a global pooling layer, and a final fully-connected classification layer; trained with the Adam optimizer and cross-entropy loss.
    \item Linear Classifiers using Random Convolutional Kernels (Rocket): Generating a large number of random
convolutional kernels to transform time series and then using the transformed features to train a linear classifier, has been shown to achieve state-of-the-art accuracy for TSC. The Rocket \citep{Rocket} method is in addition significantly faster than any other method of comparable accuracy. We use the RocketClassifier implemented in Sktime\citep{DBLP:journals/corr/abs-1909-07872}

\end{itemize}

We first check all five classification algorithms to see which is most suited.
Prior to training, we normalize both training and test sets. For datasets that do not have a validation set, we split off 20\% of the training data and use it for validation. We then fit each model on the normalized training data using default hyperparameters, and validate its performance using the validation set. 
We evaluate these five widely-used TSC algorithms, which span the spectrum from simple, interpretable white-box models to black-box deep learners. 
The accuracy on the validation set shows that Rocket emerges as the best-performing model across our benchmark. Interpretability of Rocket for TSC is a challenge because it uses thousands of random convolutional kernels making it difficult to understand which features are truly important. Hence we view Rocket as a black-box
model where interpretability is key, and focus exclusively on it in the rest of the paper, as it is a state-of-the-art TSC and interesting for applying simplifications.

\section{Experiments and Results}

It follows from the previous discussion (see Simplifications and Metrics section), where we introduced our metrics,  that for interpretability of a classifier $C_D$ we are seeking simplification algorithms that have low complexity and yet at the same time high loyalty, or rather high kappa loyalty taking into account imbalances in the distribution between classes. 

We start this section with a small experiment\footnote{Code for this and later experiment can be found at \url{https://github.com/femartip/XAI_time_series}} showing that simplifications that use a subset of the original time points as pivot points, rather than introducing pivot points that do not exist in the original time series, are more faithful and will improve trustworthiness, as the pivot points retained are often high-impact time steps. 
In the following part of this section, we then investigate which of the four simplification algorithms have the best complexity versus loyalty performance, and we will do this over a variety of characteristics of dataset, while focusing exclusively on the Rocket classifiers (see Datasets and Classifiers section).
In the last part of this section, we focus on the simplification algorithms having the best performance and show that the actual values we get for loyalty versus complexity are very good.

\subsection{Faithfulness of the simplification algorithms}


We conduct a small experiment comparing the four simplification algorithms RDP, OS, VW, BU with the two algorithms SLS (Segmented Least Squares) and PAA (Piecewise Aggregate Approximation) - that are not restricted to selecting a subset of original data points -  to see how faithful they are in retaining as pivots those data points that are most important for the classification. SLS is a Piecewise Linear Approximation, but it uses the least square method to find the straight line minimizing Euclidean distance over a segment which may lead it to create new $y$-values for the pivot points, instead of selecting subsets of the original data points. PAA  creates fixed sized windows, and replaces this area with the aggregate mean which again may lead to not choosing original data points.

It is well known that Shapley values accurately describe how important each data point is for classification of a time series, and so Shapley values are our focus. 
We measure two things: - how often the data point with maximum Shapley value is chosen as pivot, and - how many data points have higher Shapley value than the chosen pivot point of highest Shapley value. Our experiment focused on three datasets of different lengths Chinatown (length 24), SonyAIBORobotSurface1 (length 70) and ECG200 (length 96), using the Rocket classifier.
For a fair comparison we ensure that all six algorithms output approximately the same number of segments (7 out of 24 for Chinatown, 9 of 70 for Sony, and 12 of 96 for ECG200). 
For each dataset and each algorithm, we iterate over each of the 100 selected time series in the dataset and collect the data necessary for our observations, using KernelExplainer\citep{shapKernelExplainer} to compute Shapley values. 

As shown in Table \ref{tab:algo_performance_overview}, RDP achieves the highest average performance, selecting the data point of highest Shapley value 28\% of the time, and on expectation there will be 4 data points having higher Shapley value than the highest chosen. As expected, the four simplification algorithms in our study significantly outperform the two alternative methods that are not restricted to picking original data points. Indeed, SLS selects the most important data point only $10\%$ of the time, while PAA fails to do this entirely. However, it happens over 20\% of the time that SLS picks a pivot point at a time step $x$ where the $(x,y)$ data point has highest Shapley value, with SLS using some $(x,y')$ as pivot. In these cases, the average error of SLS in the $y$-coordinate is not more than $0.02$, which may often be acceptable. For PAA the similar values are not quite so good, being $10\%$ and $0.94$, respectively. We conclude that our choice of looking at simplification algorithms that retain original data points is sound when viewed from the perspective of wanting to include pivot points important for the classification.


\begin{table*}[]
\caption{ 
    Assuming $(x,y)$ has highest Shapley value, MAX SHAP gives expectation that $(x,y)$ is selected as a pivot, while CorrectX is expectation that $(x,y')$ is a pivot for any $y'$, and Dist gives expected deviation of such $y'$ from $y$, thus lower is better. Rank shows Shapley rank of the pivot with highest Shapley value, thus a low value is better. }
\begin{center}
 \begin{tabular}{|l|c|c|c|c|}
\hline
Algo & MAX SHAP $\uparrow$ & CorrectX $\uparrow$ & Dist $\downarrow$ & Rank $\downarrow$ \\
\hline
RDP & $\textbf{28.0}\%$ & $\textbf{28.0}\%$  & $0.0$  & $4.68$ \\
OS & $24.0\%$ & $24.0\%$  & $0.0$  & $\textbf{4.24}$ \\
VW & $20.0\%$ & $20.0\%$  & $0.0$  & $4.78$ \\
BU & $20.0\%$ & $20.0\%$  & $0.0$  & $4.78$ \\
\hline
SLS & $10.0\%$ & $20.0\%$  & $0.02$  & $8.0$ \\
PAA & $0.0\%$ & $10.0\%$  & $0.94$  & $10.7$ \\
\hline
\end{tabular}
 \end{center} 
    \label{tab:algo_performance_overview}
\end{table*}
\vspace{-0.3cm}

\subsection{Comparing four algorithms}

For the main experiments we  have the 4 simplification algorithms BU, OS, VW and RDP with normalized complexity parameter $\alpha_c$, and we also have a  Rocket classifier $C_D$ for each of the 40 normalized time series datasets $D$. 
The Rocket classifier achieves an average accuracy of 88\% across all datasets in the test sets. For each pair of a simplification algorithm A and a classifier $C_D$ (for a dataset $D$) we want to evaluate the interplay between loyalty and complexity, and do this as follows: 

\begin{itemize}
    \item Pick 100 instances P at random from the dataset $D$,  preserving the original fraction of instances from each class
    \item Loop over 100 values of the complexity parameter $\alpha_c$, from 0 to 1 in steps of 0.01. 
    \item For each instance $ts \in P$ and each value of $\alpha_c$, we run $A$ to get the simplification $sts=A(\alpha_c,ts)$. 
    \item We record two things: 
    \begin{itemize}
        \item $(||sts||+1)/|ts|$, to compute complexity 
        \item $C_D(ts)==C_D(sts)$, (True or False) to compute loyalty
        \end{itemize}
    \item For each value of $\alpha_c$, over all 100 $ts \in P$, we then compute the complexity, loyalty and kappa loyalty as per Definition 1:
    \begin{itemize}
      \item the average complexity in the 100 simplifications having this value of $\alpha_c$
        \item the percentage of loyal simplifications, also called agreement
        \item Cohens kappa value for the loyalty agreement (which takes into account unbalanced classes)
     
    \end{itemize}
\end{itemize}

\begin{figure}[h]

    \centering
    \includegraphics[width=12cm]{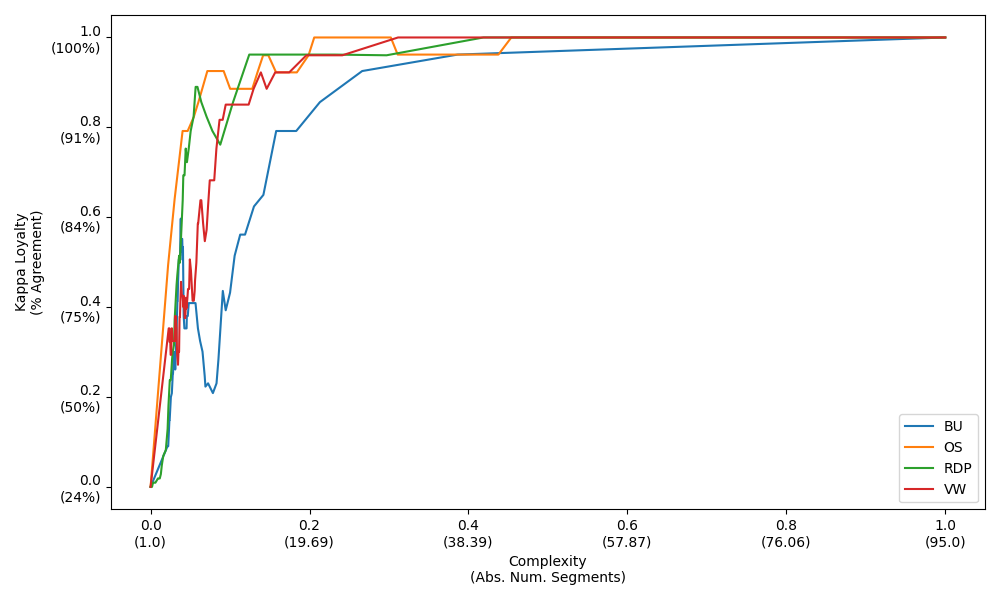}

    \caption{ECG200 dataset: Plot of Kappa loyalty (percent loyalty in parenthesis) versus Complexity (average number of segments in parenthesis) of the Rocket classifier, for all 4 simplification algorithms.}
    \label{fig:enter-label}
    \vspace{0.5cm}

\end{figure}

\begin{figure}[h]
    \centering
    \includegraphics[width=12cm]{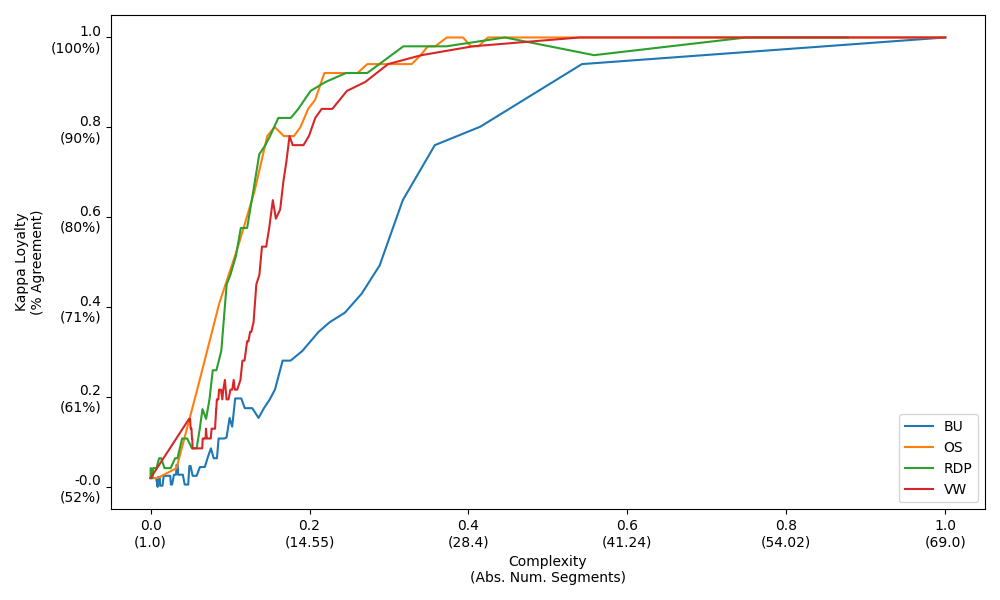}    \caption{SonyAIBORobotSurface1 dataset: Plot of Kappa loyalty (percent loyalty in parenthesis) versus Complexity (average number of segments in parenthesis) of the Rocket classifier, for all 4 simplification algorithms.}
    \label{fig:enter-label2}
    \vspace{0.5cm}
\end{figure}

  After doing the above we have, for each triple $(A,\alpha_c,C_D)$ of algorithm, complexity parameter and classifier, 100 values of kappa loyalty and complexity, one for each instance in $P$. For each algorithm and classifier pair $(A,C_D)$, to visualize the results of these 10.000 simplifications (100 instances in $P$ times 100 distinct values of $\alpha_c$)  we plot a curve.  See  Figures \ref{fig:enter-label} and \ref{fig:enter-label2} showing kappa loyalty (loyalty in parenthesis) versus complexity (number of segments in parenthesis) for all 4 simplifications algorithms on Rocket classifiers $C_D$ on the two representative datasets ECG200 and SonyAIBORobotSurface1. For both, we see that loyalty stays high even for quite low complexity values, and this is typical for most of the 40 datasets.


 To compare the performance of the four simplification algorithms, we measure the AUC - Area Under Curve -  as a value between 0 and 100, of the 40 such kappa loyalty versus complexity curves for each simplification algorithm. In  Table \ref{tab:auc} we show in the first row the average AUC for each of the four algorithms, and it is clear that OS and RDP have the best performance on average. The table also shows the average performance on classifiers for datasets of various characteristics, like Number of Classes, Stationarity, Seasonality and Entropy. We see that datasets that are Non-Stationary, Seasonal and with Low Entropy achieve higher AUC values.

\begin{table*}[h!]
\caption{AUC values for the 4 algorithms on the Rocket classifier. In the top row (Mean), we see that on average over all 40 datasets, OS achieves the highest value. The other rows display an average across datasets of varying characteristics. 
}
\centering
\begin{tabular}{|l|l|l|l|l|l|}
\hline
\textbf{Metric}              & \textbf{OS}   & \textbf{RDP}  & \textbf{BU} & \textbf{VW} & \textbf{Datasets} \\ \hline
Mean                         & \textbf{89.57} & 88.65          & 82.23        & 87.92        & 40(all)           \\ \hline \hline
Binary                       & \textbf{91.36} & 90.71          & 86.49        & 89.72        & 18                \\ \hline
Multiclass (\textgreater{}2) & \textbf{88.10} & 86.97          & 79.63        & 86.45        & 22                \\ \hline \hline
Stationary                   & \textbf{85.06} & 83.97          & 79.32        & 82.08        & 7                 \\ \hline
Non-Stationary               & \textbf{90.40} & 89.55          & 82.71        & 89.10        & 30                \\ \hline
Partially-Stationary         & \textbf{91.71}          & 90.62 & 84.22        & 89.76        & 3                 \\ \hline \hline
Seasonal                     & \textbf{95.90} & 94.73          & 90.21        & 95.52        & 8                \\ \hline
Non-Seasonal                 & \textbf{87.98} & 87.13          & 80.08        & 86.02        & 32                \\ \hline \hline
Low-Entropy                  & \textbf{91.83} & 90.01          & 84.57        & 90.57        & 14                \\ \hline
Medium-Entropy               & \textbf{88.53} & 87.59          & 79.41        & 86.09        & 12                 \\ \hline
High-Entropy                 & 88.19 & \textbf{88.20}          & 82.24        & 86.84        & 14                 \\ \hline
\end{tabular}

\label{tab:auc}
\end{table*}



We also measured performance of the four simplification algorithms at fixed loyalty values of 0.8, 0.85, 0.9, 0.95 and 1.0.  In Table \ref{tab:fix} we see that again OS has the best performance, with a resulting best average complexity of 0.15 at loyalty 0.9 and 0.35 at loyalty 1.0, which means that the time points kept in the simplification at these high loyalty values is on average $15\%$ and $35\%$. This is a very good indicator that simplifications are useful in interpretability of TSC, in particular the fact that OS simplifications keeping only 35\% of the points often have loyalty 1.0, i.e. with simplified time series not altering its classification.
In Table \ref{tab:big} we show the average number of segments of the simplifications, at various loyalty values, for both OS and RDP, for all 40 datasets.

\begin{table*}[]
\caption{Average Complexity values for the 4 algorithms on the Rocket classifiers. Over all the loyalty values we see that OS performs best and RDP second best.}
\centering
\begin{tabular}{|l|l|l|l|l|}
\hline
\textbf{Loyalty}                  & \textbf{OS}     & \textbf{RDP}    & \textbf{BU}     & \textbf{VW}     \\ \hline
\textbf{0.8}                      & \textbf{0.11}        & 0.12        & 0.18        & 0.12        \\ \hline
\textbf{0.85}                     & \textbf{0.13}        & 0.14        & 0.21        & 0.14        \\ \hline
\textbf{0.9}                      & \textbf{0.15}        & 0.17        & 0.27        & 0.17        \\ \hline
\textbf{0.95}                     & \textbf{0.19}        & 0.23        & 0.37        & 0.24        \\ \hline
\textbf{1.0}  & \textbf{0.35}   &  0.56  &   0.74    &   0.64     \\ \hline
\textbf{No Simp.}                     &     1.0    &   1.0      &    1.0     &    1.0     \\ \hline
\end{tabular}

\label{tab:fix}
\end{table*}

\subsection{A closer look at OS and RDP}

We have seen that the OS algorithm has the best average performance and also that RDP  has the best average performance out of the three faster $O(n \log n)$ algorithms. For these two algorithms, OS and RDP, we
consider the following question over a variety of dataset characteristics: for exactly what values of loyalty is the complexity low enough that the user may benefit from seeing the simplifications?

\begin{figure}[h!]
    \centering
    \includegraphics[width=10cm]{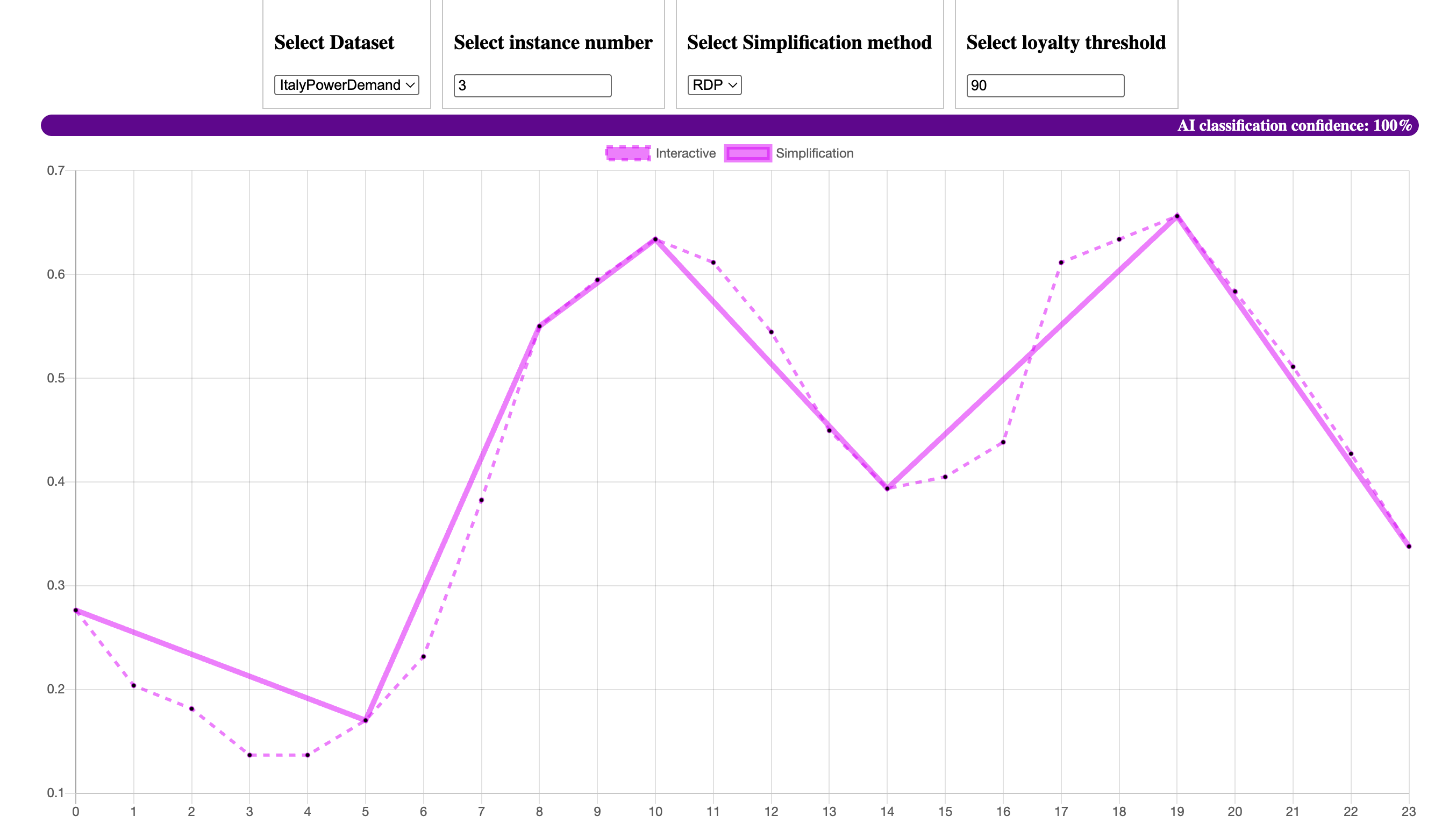}
    \caption{Screenshot of a tool where the user can select a loyalty threshold and get a simplification, using a chosen algorithm, having the fewest number of segments and expected loyalty meeting the threshold.}
    \label{fig:tool2}
    \vspace{0.4cm}
\end{figure}

Let us also mention that we have been experimenting with a tool 
for interpretability of a TSC $C_D$ using the OS and RDP simplification algorithms. See a screenshot in Figure \ref{fig:tool2}. 

\begin{figure}[h!]
    \centering
    \includegraphics[width=10cm]{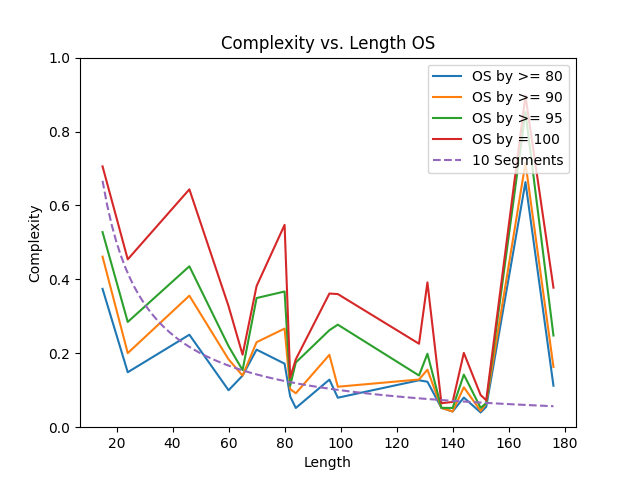}
    \caption{OS Algorithm: Complexity versus time series length for 4 loyalty values (times 100) from 80\% to 100\%, over all 40 datasets. The dotted line shows where simplifications on 10 segments would lie. Note that even with loyalty 1.0 we usually achieve complexity below 0.6. The steep value at length 166 is for the ChlorineConcentration dataset.}
    \label{fig:os}
\end{figure}

The user selects an algorithm, say RDP, and a loyalty value, say 90\%, and the tool will first find the minimum $\alpha_c$ value so that RDP with this parameter value has expected loyalty at least 90\%. 
In the next step, prototypes for the classification of dataset $D$ by $C_D$ are computed, say two sets $P_0$ and $P_1$ for $D$ being binary. 
In the final step, instead of showing these prototypes to the user, we show the simplifications $RDP(\alpha_c, ts)$, for each $ts$ in $P_0$ and $P_1$, together with the class $P_0$ or $P_1$ that it belongs to. This way the user can make his or her own mental model of how $C_D$ makes its classification. As stated above, the main question is if the number of segments in these simplifications is small enough that the user is willing to pay the price of the loyalty being only $90\%$. 
Clearly, this will depend on the length of the original time series, with shorter time series having fewer segments in their simplifications.
In Figure \ref{fig:os}
we address this question, by showing  complexity (fraction of time points kept in the simplification) across various times series lengths and how this varies as the loyalty values vary, for OS simplifications. Even for loyalty 1.0 we see that the simplifications mostly result in a complexity of less than 0.6, i.e. a $40\%$ reduction in the number of time points. This is an indicator that simplifications are useful in interpretability of TSC, and may be caused by the original time series having several consecutive data points lying on almost straight lines. Note the steep value at time series length 166, which corresponds to the dataset ChlorineConcentration.  In Table \ref{tab:big} we see that this dataset indeed has a very high number of segments for loyalty values (times 100) of 90, 95 and 100.

\begin{table*}[ht]
\vspace{0.2cm}
\centering
\caption{40 UCR datasets. The first 5 columns denote Cl. (class length), Len. (length of time series) , Stat. (stationarity), Seas. (seasonality)  and Entr. (entropy). The next 4 show average number of segments in the simplifications produced by OS, on the 100 instances chosen randomly, for loyalty values (times 100) of 90 to 100. The last 3 columns show the same for RDP.}
\begin{footnotesize}

\begin{tabular}{|l|l|l|l|l|l||r|r|r||r|r|r|}
\hline
{  \textbf{Name}}                                      & {  \hspace{-0.2cm}\textbf{Cl.}} & {  \hspace{-0.2cm}\textbf{Len.}} & { \hspace{-0.2cm} \textbf{Stat.}} & {  \hspace{-0.2cm}\textbf{Seas.}} & {  \hspace{-0.2cm}\textbf{Entr.}} & {  \textbf{\begin{tabular}[c]{@{}l@{}}\hspace{-0.2cm}OS\\90\end{tabular}}} & 
{  \textbf{\begin{tabular}[c]{@{}l@{}}\hspace{-0.2cm}OS\\95\end{tabular}}} & {  \textbf{\begin{tabular}[c]{@{}l@{}}\hspace{-0.2cm}OS\\100\end{tabular}}} &{  \textbf{\begin{tabular}[c]{@{}l@{}}\hspace{-0.2cm}\scriptsize{RDP}\\90\end{tabular}}} & {  \textbf{\begin{tabular}[c]{@{}l@{}}\hspace{-0.2cm}\scriptsize{RDP}\\95\end{tabular}}} & {  \textbf{\begin{tabular}[c]{@{}l@{}}\hspace{-0.2cm}\scriptsize{RDP}\\100\end{tabular}}} \\ \hline
Adiac & 37 & 176 & True & True & 0.27 & 28.5 & 43.4 & 66.1 & 94.6 & 134.8 & 175.0 \\ \hline

BME & 3 & 128 & False & True & 0.12 &  6.3 & 6.6 & 7.6 & 5.7 & 6.4 & 17.0 \\ \hline

CBF & 3 & 128 & False & False & 0.83 &  32.8 & 35.0 & 43.1 & 20.4 & 23.3 & 32.9 \\ \hline

Chinatown & 2 & 24 & False & False & 0.23 &   2.3 & 2.4 & 8.3 & 4.1 & 4.6 & 12.3 \\ \hline

ChlorineConct & 3 & 166 & True & False & 0.78 & 109.4 & 118.0 & 148.0 & 127.9 & 146.4 & 165.0 \\ \hline

Crop & 24 & 46 & False & False & 0.37 & 15.7 & 19.2 & 29.0 & 15.9 & 24.5 & 41.2 \\ \hline

DistalPhxOutlAge & 3 & 80 & False & False & 0.27 & 10.8 & 13.7 & 58.7 & 8.7 & 17.2 & 70.9 \\ \hline

DistalPhxOutlCor& 2 & 80 & False & False & 0.30 & 9.7 & 10.5 & 59.8 & 9.9 & 13.6 & 72.0 \\ \hline

DistalPhxTW & 6 & 80 & False & False & 0.27 & 18.1 & 22.6 & 36.5 & 17.8 & 28.9 & 69.8 \\ \hline

ECG200 & 2 & 96 & False & False & 0.46 &  4.6 & 7.0 & 20.4 & 5.2 & 6.3 & 40.4 \\ \hline

ECG5000 & 5 & 140 & False & False & 0.24 &  4.7 & 5.9 & 9.5 & 5.8 & 6.2 & 29.9 \\ \hline

ECGFiveDays & 2 & 136 & Partial & False & 0.20 &  6.1 & 6.6 & 8.8 & 6.6 & 6.9 & 7.8 \\ \hline

ElectricDevices & 7 & 96 & Partial & False & 0.29 &  25.8 & 31.4 & 48.4 & 28.0 & 37.7 & 50.4 \\ \hline

FaceAll & 14 & 131 & True & False & 0.58 &   21.7 & 30.6 & 73.5 & 20.9 & 30.0 & 129.3 \\ \hline

FacesUCR & 14 & 131 & True & False & 0.55 & 16.3 & 18.7 & 28.3 & 19.4 & 22.1 & 26.4 \\ \hline

GunPoint & 2 & 150 & False & True & 0.11 & 5.3 & 6.0 & 7.8 & 6.2 & 7.2 & 10.1 \\ \hline

GunPointAgeSpan & 2 & 150 & False & True & 0.13 & 4.6 & 5.4 & 14.9 & 4.9 & 8.1 & 148.9 \\ \hline

GunPointMalVsFem & 2 & 150 & False & True & 0.12 &  5.1 & 5.5 & 8.9 & 4.6 & 5.8 & 6.6 \\ \hline

GunPointOldVsYou & 2 & 150 & False & True & 0.12 &  5.4 & 6.4 & 8.7 & 6.7 & 7.7 & 13.9 \\ \hline

ItalyPowerDemand & 2 & 24 & False & False & 0.24 & 4.4 & 4.9 & 12.6 & 4.2 & 5.2 & 14.4 \\ \hline

MedicalImages & 10 & 99 & False & False & 0.18 & 9.0 & 12.6 & 35.3 & 17.8 & 42.0 & 98.0 \\ \hline

MiddlePhxOutlAge & 3 & 80 & False & False & 0.23 & 17.5 & 35.3 & 49.0 & 29.6 & 35.9 & 77.9 \\ \hline

MiddlePhxOutlCor & 2 & 80 & False & False & 0.24 & 17.8 & 19.1 & 28.8 & 17.8 & 22.9 & 77.3 \\ \hline

MiddlePhxTW & 6 & 80 & False & False & 0.25 & 26.6 & 31.6 & 64.1 & 28.1 & 35.4 & 71.2 \\ \hline

MoteStrain & 2 & 84 & False & False & 0.31 & 4.0 & 7.6 & 15.2 & 6.6 & 7.8 & 82.2 \\ \hline

PhalangesOutlsCor & 2 & 80 & False & False & 0.25 & 11.8 & 13.7 & 24.1 & 16.2 & 19.4 & 30.7 \\ \hline

Plane & 7 & 144 & False & False & 0.35 & 8.2 & 8.6 & 9.0 & 9.0 & 9.3 & 14.9 \\ \hline

PowerCons & 2 & 144 & False & False & 0.36 & 12.4 & 20.2 & 48.5 & 9.5 & 16.7 & 133.2 \\ \hline

ProxPhxOutlAge& 3 & 80 & Partial & False & 0.24 & 9.6 & 17.8 & 39.6 & 13.6 & 21.8 & 65.2 \\ \hline

ProxPhxOutlCor & 2 & 80 & False & False & 0.23 & 16.2 & 22.3 & 41.1 & 20.5 & 28.2 & 63.3 \\ \hline

ProxPhxTW & 6 & 80 & False & False & 0.23 & 10.8 & 16.0 & 30.8 & 14.7 & 25.6 & 67.8 \\ \hline

SmoothSubspace & 3 & 15 & False & False & 0.04 & 5.5 & 6.7 & 9.9 & 6.0 & 7.0 & 10.9 \\ \hline

SonyAIBORobotS1 & 2 & 70 & True & False & 0.35 & 11.6 & 15.6 & 26.4 & 11.6 & 16.0 & 31.3 \\ \hline

SonyAIBORobotS2 & 2 & 65 & True & False & 0.38 & 7.7 & 8.7 & 12.6 & 8.1 & 9.2 & 22.6 \\ \hline

SwedishLeaf & 15 & 128 & False & True & 0.42 & 11.2 & 15.5 & 53.0 & 11.1 & 17.1 & 43.1 \\ \hline

SyntheticControl & 6 & 60 & False & False & 0.46 & 10.1 & 12.4 & 19.3 & 8.6 & 9.7 & 20.6 \\ \hline

TwoLeadECG & 2 & 82 & False & False & 0.33 & 6.7 & 8.4 & 10.9 & 7.9 & 8.5 & 9.4 \\ \hline

TwoPatterns & 4 & 128 & True & False & 0.70 &  9.7 & 10.4 & 11.0 & 7.4 & 7.9 & 12.5 \\ \hline

UMD & 3 & 150 & False & True & 0.12 & 7.0 & 7.9 & 24.5 & 7.3 & 9.8 & 34.7 \\ \hline

Wafer & 2 & 152 & False & False & 0.13 & 7.4 & 8.1 & 11.0 & 8.4 & 9.0 & 14.3 \\ \hline

\end{tabular}
\end{footnotesize} 
\label{tab:big}
\vspace{0.3cm}
\end{table*}

\section{End-user evaluation with forward simulation}

As time series are notoriously difficult for human users, most of the human evaluations of XAI methods done so far have been qualitative and involving domain experts. Our goal in this work has been to add the new dimension of user evaluations involving end-users, i.e. non-experts, on the use of simplifications for interpretability of TSCs. 
In this section we present an end-user evaluation with forward simulation, i.e. where participants will be trained on prototypes from each of the TSC classes, and then tested by being asked to guess classes of random instances. We first present the 4 datasets chosen for the evaluation, discuss the format of the training and testing stages, and give the results of the evaluation.

\begin{table}[t]
\centering
\caption{Summary of the datasets employed in the evaluation. For each dataset, we report the number of classes, the length, the test accuracy (wrt Ground Truth) obtained by the Rocket model, and the selected loyalty thresholds Loy and number of segments ($\#$Segs) corresponding to Knee Low and Knee High.}
\label{tab:data}
\begin{tabular}{|l|c|c|c|c|c|}
\toprule
\textbf{Dataset} &  \textbf{Cl.} & Length & \textbf{Rocket acc} & \textbf{ K.Low:Loy($\#$Seg)} & \textbf{K.High:Loy($\#$Seg)} \\
\midrule
Chinatown & 2 &  24 & 98 & 0.72(1.0)  & 0.92(2.4)  \\
ECG200    & 2 &  96 &  96  &   0.82(3.1)   &   0.97(7.7)  \\
SonyAIBORobotSurface1 & 2 &  70 & 90  &  0.90(11.6)   &  0.96(15.9)  \\
UMD    & 3 & 150&94   &  0.95(7.9)  &  0.99(13.6)  \\
\bottomrule
\end{tabular}
\end{table}

Since our survey participants are not domain experts and time series are notoriously non-intuitive to humans we  want univariate datasets with a binary or ternary classification. Moreover, we want to cover a good spread of characteristics, also regarding Stationarity, Seasonality and Entropy. We chose the four datasets in Table \ref{tab:data}. The majority of the 40 UCR datasets are neither Stationary nor Seasonal, and both Chinatown and ECG200 belong here, with the first having low entropy and the second having high entropy. The chosen Sony dataset is Stationary but not Seasonal, whereas UMD is the opposite. 
Further details of these datasets can be found in Table \ref{tab:data}. 
We also include the accuracy obtained by the Rocket TSC on the test set. For these datasets, we used the training/test split defined in the repository.

\begin{figure}
    \centering
    \includegraphics[width=0.5\linewidth]{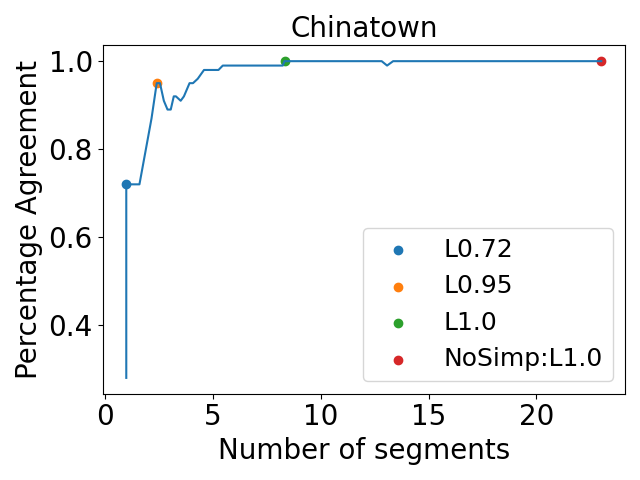}
    \caption{Complexity-loyalty curve for Chinatown using the OS algorithm. Four points have been manually selected, covering: the original dataset without simplification, the point of lowest complexity for a loyalty of 1.0, one point we call High Knee, and one point we call Low Knee.}
    \label{fig:auc_knee_survey}
\end{figure}

We also chose four simplification levels for the survey, as follows. Firstly, we are comparing against original time series, so we need a level with No simplification. Secondly, we use the simplification with lowest complexity that still achieves max loyalty of 1.0, and call this L1.0.
Thirdly, looking for a sweet spot of loyalty versus complexity we inspect the AUC curves and pick points where the second derivative seems to change sign. The highest such point we call Knee High, and the second highest is called Knee Low. 
As an example, Figure \ref{fig:auc_knee_survey} shows the complexity-loyalty curve for Chinatown using the OS algorithm with the 4 chosen simplification levels highlighted. 
These loyalty values are also given in  Table \ref{tab:data}.

We thus have 16 different user configurations, 4 datasets by 4 simplification levels, as shown in Table \ref{tab:simpl_simulatability}. Each participant will be shown 4 user configurations, ensuring that these are from 4 distinct datasets and 4 distinct simplification values, i.e. with 4 participant groups forming a 4 by 4 Latin square. For each of these 4 configurations the participant is shown one page with 3 prototypes per class, and at the same time one page with 10 random test instances, see Figures \ref{fig:example_human_survey_train} and \ref{fig:example_human_survey_test}  in the Appendix for an example. Prototypes are selected by applying KMedoids with Dynamic Time Wrapping (DTW) and obtaining the medioids of each class \citep{holder2023clustering}.

It is important to note the following. For TSC $C$, simplification algorithm $A_{\alpha}$ and an original time series $ts$ taken from dataset $D$, when we show the user a prototype or test instance $A_{\alpha}(ts)$ we consider it to be in class $C(ts)$. The loyalty $\lambda$ of the triple $(C, A_{\alpha}, D)$ is the probability that $C(A_{\alpha}(ts))=C(ts)$. Thus, when $\lambda < 1.0$ those two classifications can differ which could lead the user to misinterpret properties of the classifier on original time series, and we will take this into account when evaluating our results.

\subsection{Evaluation results}

In total, we had 19 voluntary participants, who were students in a university-level informatics course, none of whom received compensation. The participants were presented with the survey and freely chose to participate. The participants were randomly divided into the 4 survey groups. They were asked to answer 10 test questions on each of 4 configurations. We ensured that participants spent at least 15 minutes on the complete survey. 
In Table \ref{tab:simpl_simulatability} we see the absolute accuracy, i.e. the percentage of correct test answers, on each of the 16 configurations.

\begin{table*}[h!]
\centering
\caption{Evaluation Results. Mean human accuracies (in \%) for each dataset and simplification level. Values are given as mean $\pm$ half of the 95 \% bootstrap confidence interval (CI) across participants.}
\begin{tabular}{|l|c|c|c|c|}
\toprule
\textbf{Dataset} & \textbf{Knee Low} & \textbf{Knee High} & \textbf{L1.0} & \textbf{No Simp.} \\
\midrule
Chinatown              & 78.0\,\%\,$\pm$\,6.0 & 82.5\,\%\,$\pm$\,3.8  & 76.0\,\%\,$\pm$\,15.0 & 66.0\,\%\,$\pm$\,14.0\\
ECG200                 & 68.0\,\%\,$\pm$\,12.0 & 72.0\,\%\,$\pm$\,12.0 & 72.0\,\%\,$\pm$\,8.0  & 72.5\,\%\,$\pm$\,3.8  \\
SonyAIBORobotSurface1  & 67.5\,\%\,$\pm$\,3.0  & 72.0\,\%\,$\pm$\,10.0 & 76.0\,\%\,$\pm$\,15.0 & 78.0\,\%\,$\pm$\,10.0 \\
UMD                    & 98.0\,\%\,$\pm$\,3.0  & 100.0\,\%\,$\pm$\,0.0 & 100.0\,\%\,$\pm$\,0.0 & 96.0\,\%\,$\pm$\,6.0  \\
\bottomrule
\end{tabular}
\label{tab:simpl_simulatability}
\end{table*}

As discussed earlier, our goal is to teach the patterns discovered by the TSC $C$ in classifying the \emph{original time series}. However, when we consider the class label of an original time series attached to a \emph{simplified time series} with loyalty $\lambda < 1$ then what the user learns does not fully carry over to the original. For example, if $\lambda=0.8$ then there is 80\% chance that the simplified instance $A(ts)$ is classified the same as the original instance $ts$, so whatever the user learns has an $80\%$ chance of being useful for understanding the patterns used by $C$ in classifying the original time series. Hence we need to multiply the absolute user accuracies by the loyalty values to obtain the correct loyalty-adjusted accuracy with regards to the goal of understanding the TSC on the original time series.


\begin{table*}[h!]
\centering
\caption{Evaluation results. Mean loyalty-adjusted human accuracies (in \%) for each dataset and simplification level.  These are achieved by multiplying absolute accuracy from Table \ref{tab:simpl_simulatability} with loyalty from Table \ref{tab:data}.}
\begin{tabular}{|l|c|c|c|c|}
\toprule
\textbf{Dataset} & \textbf{Knee Low} & \textbf{Knee High} & \textbf{L1.0} & \textbf{No Simp.} \\
\midrule
Chinatown              & 56.6  & 75.9  & \bf{76.0}  & 66.0  \\
ECG200                 & 55.8  & 69.8  & 72.0 & \bf{72.5}  \\
SonyAIBORobotSurface1  & 60.8   & 69.1  & 76.0  & \bf{78.0}  \\
UMD                    & 93.1 & 99.0   & \bf{100.0}  & 96.0  \\
\bottomrule
\end{tabular}
\label{tab:mod_simulatability}
\end{table*}

Table \ref{tab:mod_simulatability} shows the loyalty-adjusted accuracies of the human subjects, on the 16 configurations, with values in bold showing the max scores. For Chinatown simplifications seem to be useful, either at the Knee High loyalty value 0.92, or at L1.0 loyalty value 1.0.
 However, for the other three datasets ECG200, SonyAIBORobotSurface1 and UMD, simplifications do not seem to help. In the next section we reason carefully about why simplifications help or not for these four datasets, and use this as a starting point for a general discussion involving also other datasets among the 40 covered.

\section{A framework for evaluating the utility of simplifications}

\begin{figure}[h]
\centering
\includegraphics[scale=0.23]{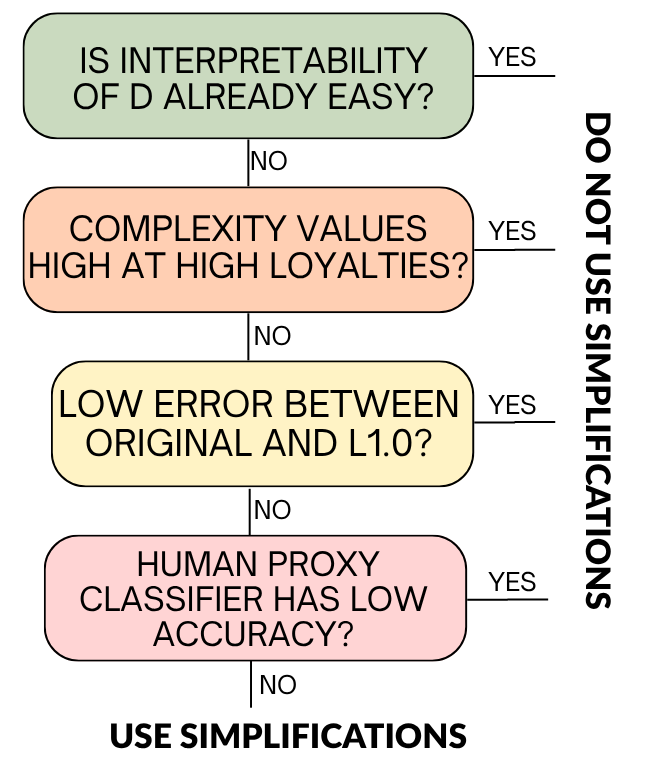}
\caption{A flowchart to estimate if a certain simplification algorithm $A$ over various parameter values $\alpha$ will improve the interpretability of a TSC $C$ on a dataset $D$.}
\label{fig:flowchart}
\end{figure}

In this section we present a framework giving a rule-of-thumb to decide if a simplification algorithm $A$ (over various $\alpha$ values) is useful for interpretability of a TSC $C$ on a dataset $D$.
The four datasets covered in our user study, together with other selected datasets among the 40 investigated, will act as instructive in explaining our reasoning that leads to the flowchart presented in Figure \ref{fig:flowchart}.

\begin{itemize}
    \item UMD dataset: Human accuracy on the original time series (No simpl) is a very high value of $96\%$ and so we cannot hope that simplifications improves this by much. Figures \ref{fig:example_human_survey_train} and \ref{fig:example_human_survey_test} in the Appendix shows the user survey for this configuration and it is indeed easy for the human eye to see the patterns used by the classifier. This observation is generalized and forms the first box in the flowchart in Figure \ref{fig:flowchart}.
    \item ChlorineConcentration dataset: This dataset is responsible for the steep value in Figure \ref{fig:os}. Also table \ref{tab:big} and also Figure \ref{fig:CC} show that as we decrease
    \begin{figure}
    \centering
    \includegraphics[width=0.7\linewidth]{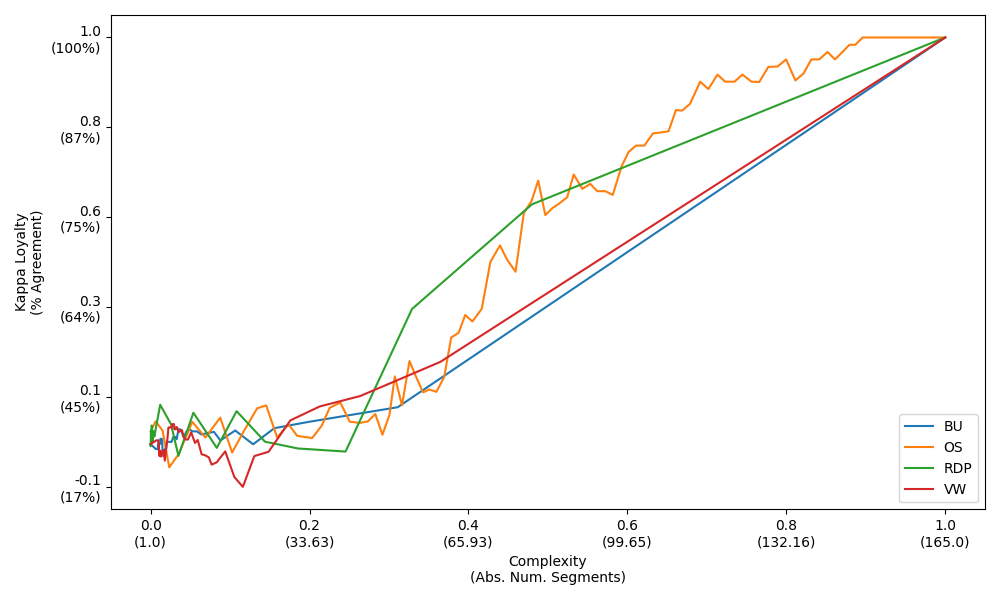}
    \caption{ChlorineConcentration: Plot of kappa loyalty versus Complexity of the Rocket classifier, for all 4 simplifcation algorithms.}
    \label{fig:CC}
\end{figure}
    loyalty the complexity values remain high. In that case we do not expect loyalty-adjusted accuracies to be better for simplifications than for No simplifications, as few data points have been removed. This observation is generalized in the second box of Figure \ref{fig:flowchart}.
    \item ECG200 dataset: Figure \ref{fig:ECG200_low_error} shows a prototype time series from the survey superimposed with its L1.0 simplification. 
\begin{figure}
    \centering
    \includegraphics[width=0.5\linewidth]{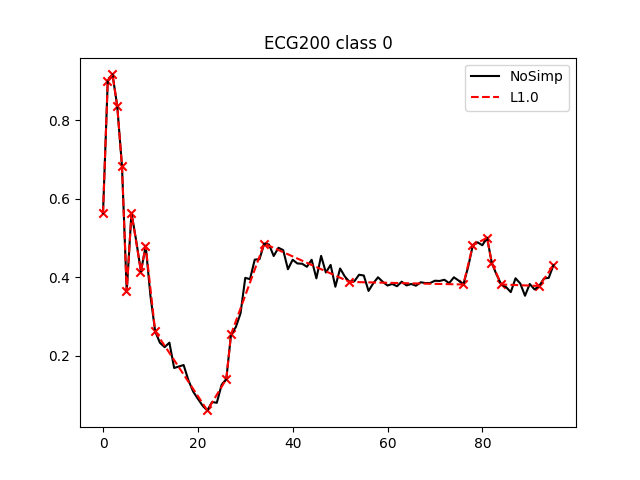}
    \caption{Figure showing the overlap between NoSimplification in black and the simplification at L1.0 in red. This instance is one of three prototypes from class 0 of ECG200 shown to the participants in the user survey, with some users seeing the black and other users seeing the red}
    \label{fig:ECG200_low_error}
\end{figure}
    Even though the latter has retained only 22 of the original 96 data points, the Euclidean distance (error) between the two is very low. This means that a human user will probably not do much better on No Simpl than on L1.0, as the human eye in any case will tend to straighten out a series of consecutive almost collinear points.
This observation is generalized in the third box of Figure \ref{fig:flowchart}.
\item SonyAIBORobotSurface1 dataset: This dataset passes the first three boxes in the flowchart, but the human accuracy levels in Tables \ref{tab:simpl_simulatability} and \ref{tab:mod_simulatability} are highest for No simplification. Could we have discovered this was likely to happen without doing a human user study? Yes, we could train a simple classifier (as proxy for  human user) on the exact same task facing the human user in the various configurations and see how well it does. Table \ref{tab:TSC_model_of_human} shows results of a logistic regressor on the 16 configurations in the survey. Note the chosen Sony dataset is the only one of consistently low accuracy. We  assume a human would also have problems and this observation is generalized in the fourth box of Figure \ref{fig:flowchart}.
\item Chinatown dataset: It passes all the tests mentioned above and therefore our rule-of-thumb tells us that the simplifications are probably useful for human interpretability.
\end{itemize}

\begin{table*}[htbp]
  \centering
    \caption{Human Proxy Classifier. Accuracy/loyalty-adjusted accuracy for logistic regressor trained on the same task as human participants face in the survey. Note that we expand training and testing sets for the logistic regressor compared to the human survey.}
  \begin{tabular}{|l|c|c|c|c|}
\hline
Dataset & Knee Low & Knee High  & L1.0 & NoSimp \\
\hline
Chinatown & $100\%$ / $72\%$ & $87\%$ / $82\%$ & $93\%$ / $93\%$ & $93\%$ / $93\%$\\
ECG200 & $80\%$ / $66\%$ & $77\%$ / $74\%$ & $80\%$ / $80\%$ & $80\%$ / $80\%$\\
SonyAIBORobotSurface1 & $50\%$ / $45\%$ & $50\%$ / $48\%$ & $50\%$ / $50\%$ & $50\%$ / $50\%$\\
UMD & $83\%$ / $79\%$ & $83\%$ / $82\%$ & $83\%$ / $83\%$ & $83\%$ / $83\%$\\
\hline
\end{tabular}

  \label{tab:TSC_model_of_human}
\end{table*}

Let us remark right away that for simplicity of presentation purposes we have made the above discussion and the flowchart in Figure \ref{fig:flowchart} somewhat simpler than reality dictates. Rather then clearcut  Yes/No answers there will sometimes be propensities. For a concrete triple of simplification algorithm $A$, TSC $C$ and dataset $D$, these propensities may or may not add up to a recommendation for using the simplifications in interpretation.

\section{Conclusion}

We have introduced metrics like complexity and loyalty, see Definition 1, to evaluate the use of simplification algorithms in interpretability of Time Series Classifiers. Human perceptual studies \citep{cleveland1984graphical,heer2010crowdsourcing} show that people read line‐slope and point‐position with near-bar accuracy, whereas decoding symbols or averages is slower and less precise, and this supports our choice of using simplifications. We have applied these metrics to evaluate four simplification algorithms on Rocket classifiers for 40 datasets from the UCR repository \citep{UCR}. The theoretical results show that the OS (Optimal Simplification) and RDP (Ramer-Douglas-Peuker) algorithms have the best performance\footnote{Note that the BU (Bottom-Up) algorithm was designed to minimise the number of segments that go over more than a single time step. However, for the purpose of the comparisons in this paper we needed to count also for BU the segments on a single time step. This explains why BU performs worse than the other algorithms. In future work we would like to compare BU to other algorithms sharing its simplification criterium.}. The OS algorithm has time complexity cubic in the length $n$ of the original time series, and hence the $O(n logn)$ RDP algorithm would be preferable for very long time series. 

Our results contribute to an understanding of when simplifications can be useful for interpretability of TSC. We performed a practical user evaluation with forward simulation.  We used our findings to arrive at a flowchart that employs the introduced metrics of complexity and loyalty to estimate whether simplifications would be useful for interpretability in a given situation. 




\bibliography{biblio}   

\newpage

\begin{appendix}

\section{Appendix. End-user evaluation}\label{secA1}

\begin{figure}[b]
    \centering
    \includegraphics[width=0.9\linewidth]{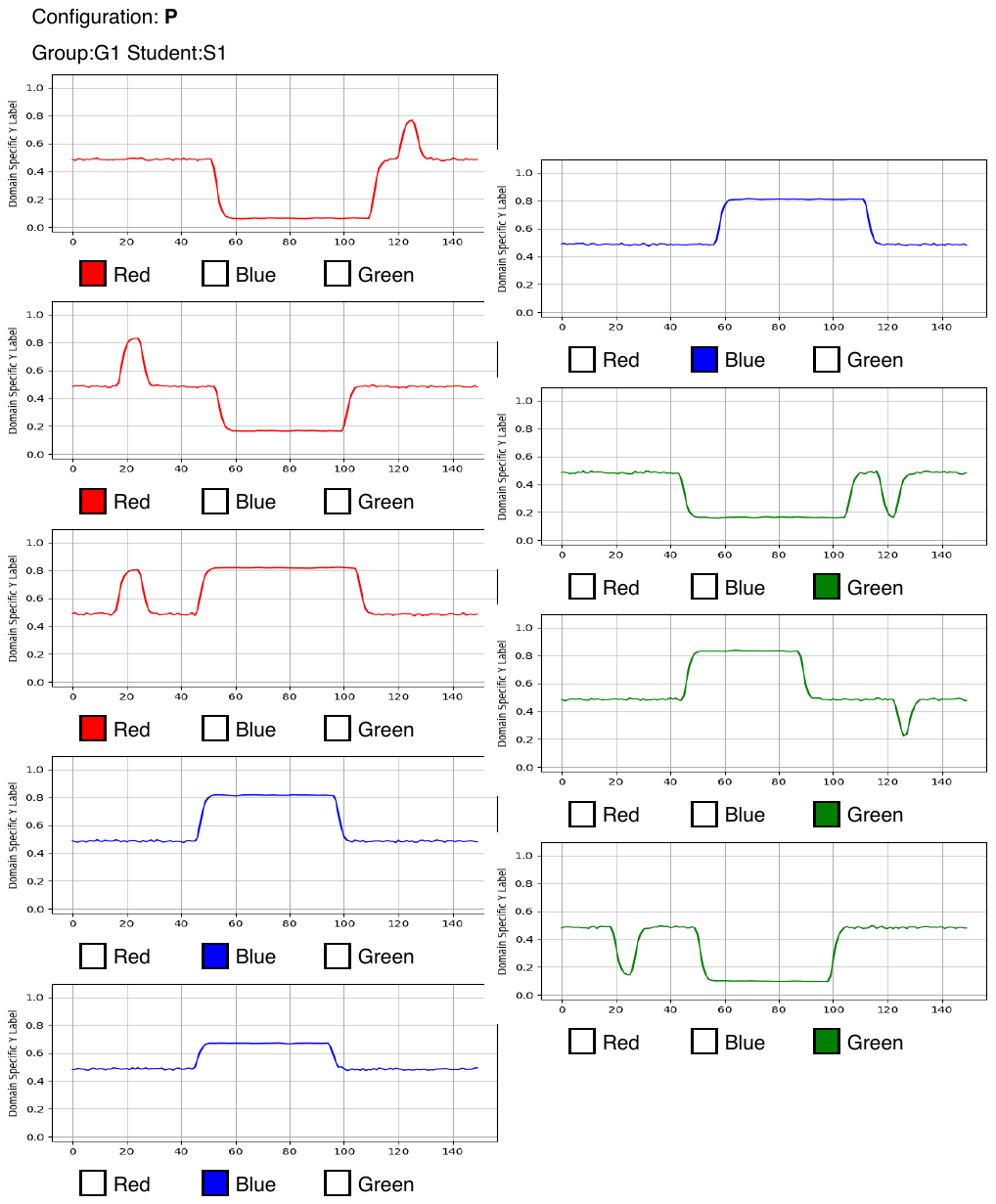}
    \caption{UMD dataset prototypes, No Simplification. Example of the layout presented in the user survey. The six sub-figures correspond to the labelled prototypes used as "training" instances for the subjects. }
    \label{fig:example_human_survey_train}
\end{figure}

\begin{figure}
    \centering
    \includegraphics[width=0.9\linewidth]{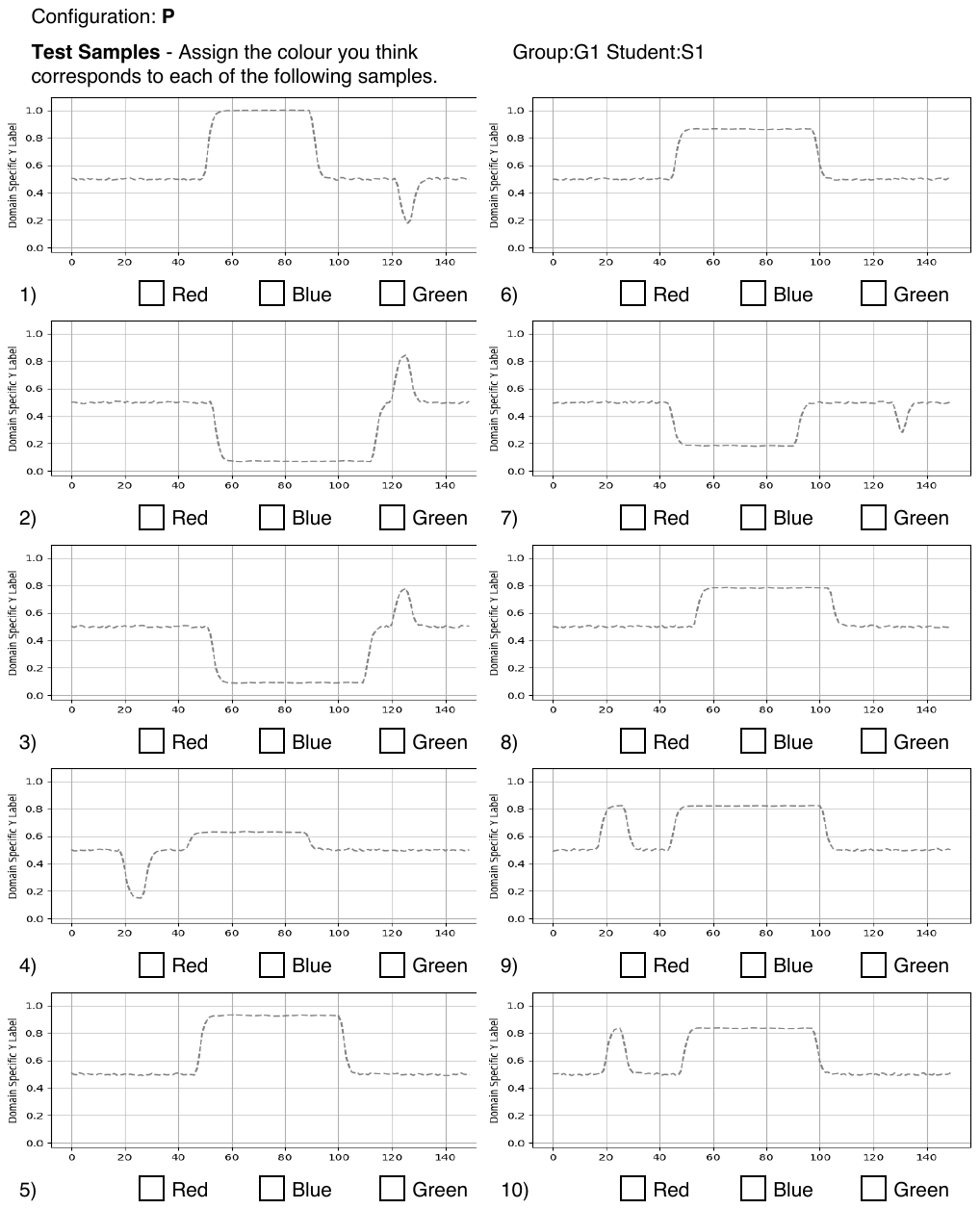}
    \caption{UMD dataset test instances, No Simplification. Example of the layout presented in the user survey. The ten sub-figures show the new, unlabelled instances they were asked to classify. For the interested reader we list the correct class classification of the test instances: 1:Green, 2:Red, 3:Red, 4: Green, 5:Blue, 6:Blue, 7:Green, 8:Blue, 9:Red, 10:Red}
    \label{fig:example_human_survey_test}
\end{figure}

Survey conducted with humans. Figure \ref{fig:example_human_survey_train} shows instances with labels used for the users as reference, and Figure \ref{fig:example_human_survey_test} shows unlabelled instances that the users were asked to classify.


\end{appendix}


\end{document}